\documentclass[final]{cvpr}

\usepackage[pagebackref=true,breaklinks=true,colorlinks,bookmarks=false]{hyperref}

\usepackage{times}
\usepackage{graphicx}
\usepackage{amsmath,amssymb,amsthm}
\usepackage{algorithmic}
\usepackage{acronym}
\usepackage{enumitem}
\usepackage{balance}
\usepackage{setspace}
\usepackage[dvipsnames]{xcolor}
\usepackage[capitalise]{cleveref}
\usepackage{tabularx,multirow,array,makecell}
\usepackage{overpic,wrapfig}

\newcommand{\norm}[1]{\left\Vert #1 \right\Vert}

\makeatletter
\renewcommand{\paragraph}{%
  \@startsection{paragraph}{4}%
  {\z@}{0ex \@plus 0ex \@minus 0ex}{-1em}%
  {\hskip\parindent\normalfont\normalsize\bfseries}%
}
\makeatother

\graphicspath{{figures/}}

\crefname{algocf}{Alg.}{Algs.}
\Crefname{algocf}{Algorithm}{Algorithm}

\definecolor{gblue}{HTML}{4285F4}
\definecolor{gred}{HTML}{DB4437}

\acrodef{rpm}[RPM]{Raven's Progressive Matrices}
\acrodef{wren}[WReN]{Wild Relational Network}
\acrodef{ai}[AI]{Artificial Intelligence}
\acrodef{acre}[ACRE]{Abstract Causal REasoning}
\acrodef{ood}[O.O.D.]{Out-Of-Distribution}
\acrodef{iq}[IQ]{Intelligence Quotient}
\acrodef{iid}[I.I.D.]{Independent and Identically Distributed}
\acrodef{dag}[DAG]{Directed Acyclic Graph}
\acrodef{rw}[RW]{Rescorla--Wagner}
\acrodef{sem}[SEM]{Structural Equation Model}
\acrodef{mlp}[MLP]{Multilayer Perceptron}

\def\cvprPaperID{xxxx} 
\def\confYear{CVPR 2021}

\begin{document}

\title{ACRE: \underline{A}bstract \underline{C}ausal \underline{RE}asoning Beyond Covariation}

\author{Chi Zhang \qquad Baoxiong Jia \qquad Mark Edmonds \qquad Song-Chun Zhu \qquad Yixin Zhu \\
UCLA Center for Vision, Cognition, Learning, and Autonomy \\
{\tt\small \{chi.zhang,baoxiongjia,markedmonds\}@ucla.edu, sczhu@stat.ucla.edu, yixin.zhu@ucla.edu}
}

\maketitle
\thispagestyle{empty}
\pagestyle{empty}

\begin{abstract}
Causal induction, \ie, identifying unobservable mechanisms that lead to the observable relations among variables, has played a pivotal role in modern scientific discovery, especially in scenarios with only sparse and limited data. Humans, even young toddlers, can induce causal relationships surprisingly well in various settings despite its notorious difficulty. However, in contrast to the commonplace trait of human cognition is the lack of a diagnostic benchmark to measure causal induction for modern \ac{ai} systems. Therefore, in this work, we introduce the \ac{acre} dataset for systematic evaluation of current vision systems in causal induction. Motivated by the stream of research on causal discovery in \emph{Blicket} experiments, we query a visual reasoning system with the following four types of questions in either an independent scenario or an interventional scenario: \emph{direct}, \emph{indirect}, \emph{screening-off}, and \emph{backward-blocking}, intentionally going beyond the simple strategy of inducing causal relationships by covariation. By analyzing visual reasoning architectures on this testbed, we notice that pure neural models tend towards an associative strategy under their chance-level performance, whereas neuro-symbolic combinations struggle in backward-blocking reasoning. These deficiencies call for future research in models with a more comprehensive capability of causal induction.
\end{abstract}

\section{Introduction}

\begin{quote}
    ``There is something fascinating about science. One gets such wholesale returns of conjecture out of such a trifling investment of fact.'' 
    
    \hfill --- Mark Twain~\cite{twain1984life}
\end{quote}

The history of scientific discovery is full of intriguing anecdotes. Mr. Twain is accurate in summarizing how influential science theories are distilled from sparse and limited investments. From only three observations, Edmond Halley precisely predicted the orbit of the Halley comet and its next visit, which he did not live to see. From a few cathode rays, Joseph Thomson proved and derived the existence of electrons. From merely crossbreeding of pea plants, Gregor Mendel established the laws of Mendelian inheritance much beyond pea plants. Out of many other possible conjectures, pioneering scientists picked the most plausible ones.

\begin{figure}[t!]
    \centering
    \includegraphics[width=\linewidth]{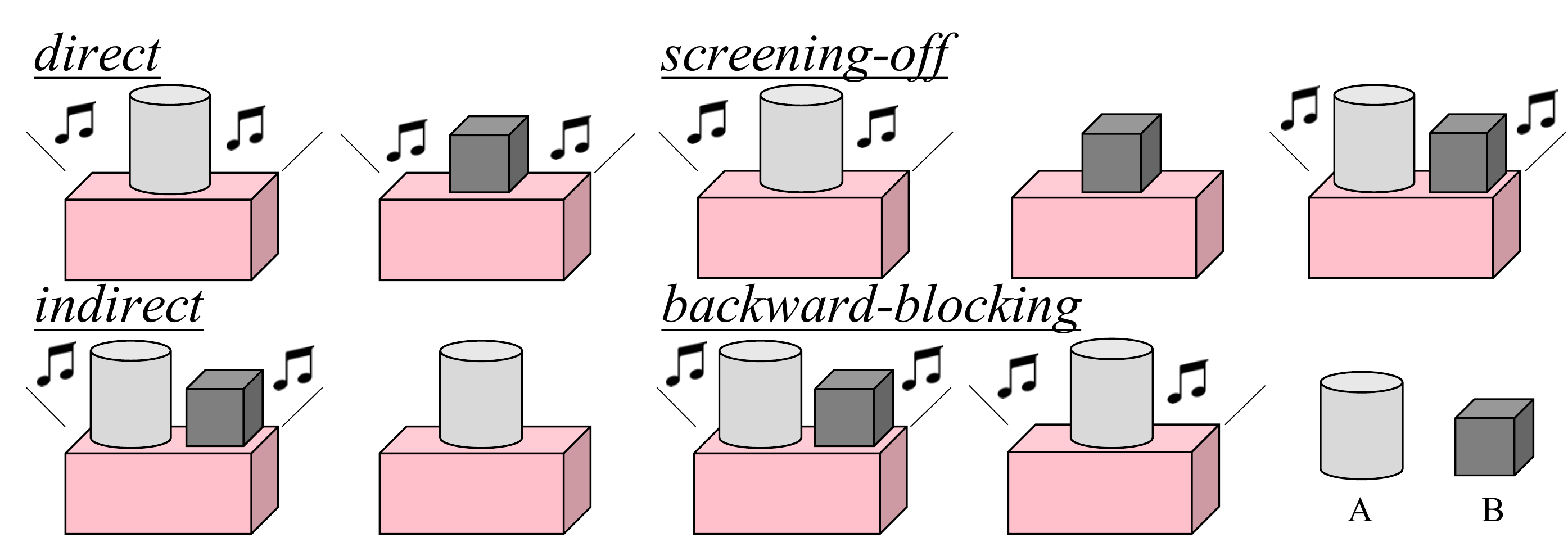}
    \caption{Abstract causal reasoning tasks administered to human participants~\cite{gopnik2001causal,sobel2004children}. The Blicket machine possesses various activation patterns in these four cases. One needs to discover the hidden causal relations to answer two types of questions: whether object A / B is a Blicket, and how to make the machine stop / go.}
    \label{fig:intro}
\end{figure}

The above examples of causal induction are only a few acclaimed cases of omnipresent causal reasoning scenarios in science history and our daily life. In fact, despite the notorious complexity in causal discovery, humans, even young toddlers, can felicitously identify and, sometimes, intervene in the unobservable mechanisms from only a trifling number of samples of observable events~\cite{gopnik2012scientific,schulz2007learning}. 

This captivating commonplace trait of human cognition and its paramount connection to human learning mechanism motivate us to ask a counterpart question for modern \acf{ai} systems: 
\begin{quote}
    \textit{At what level do current visual reasoning systems induce causal relationships?}
\end{quote}

To answer this question, we propose the \acf{acre} dataset. \ac{acre} is inspired by the established stream of research on \textit{Blicket} detection originally administered to young toddlers~\cite{cook2011science,gopnik2012scientific,gopnik2004theory,gopnik2000detecting,gopnik2001causal,gopnik2012reconstructing,kushnir2007conditional,lucas2014children,meltzoff2012learning,schulz2007learning,sobel2006blickets,sobel2004children,walker2014toddlers,walker2017discriminating}. The original experiments designed by Gopnik and Sobel~\cite{gopnik2000detecting} introduced a novel setup for investigating children's ability of causal induction, in which children were given a special machine referred to as ``Blicket detector.'' Its underlying mechanism is intuitive: A Blicket detector would activate, lighting up and making noise, when a ``Blicket'' was put on it. The experimenter demonstrated a series of trials to participants by placing various (combinations of) objects on the Blicket detector and showing whether the detector was activated or not. At length, the participants were asked which object is a Blicket and how to make an (in)activated Blicket machine stop (go).

This line of work's intricate nature lies in how the context and query were designed to test abstract causal reasoning beyond the simple strategy of covariation; see an illustration in \cref{fig:intro}. As a base test on causal discovery by covariation, Sobel \etal~\cite{sobel2004children} show that children can correctly associate cause and effect using \textit{direct} evidence. They also show that with only \textit{indirect} evidence asserting the Blicketness of object B, children still made accurate predictions~\cite{gopnik2001causal}. However, one must go beyond the simple covariation strategy to discover the hidden causal relations in the \textit{screening-off} case and the \textit{backward-blocking} case. Specifically, in the screening-off setting (\cref{fig:intro} Top), object B (non-Blicket) is screened-off by A (Blicket) from probabilistically activating the machine~\cite{gopnik2001causal}. The backward-blocking setting (\cref{fig:intro} Bottom) is even more intriguing as object B, not independently tested, has undetermined Blicketness despite the fact that every appearance of it is associated with an activated machine~\cite{sobel2004children}. See \cref{sec:acre} for details and the supplementary for a symbolic summary.

The proposed \ac{acre} dataset is built following a similar querying manner in the Blicket experiments to study how well existing visual reasoning systems can learn to derive ample causal information from scarce observation. In particular, inspired by the recent endeavors of visual reasoning in controlled environments~\cite{girdhar2020cater,johnson2017clevr,yi2020clevrer}, we adopt the CLEVR universe~\cite{johnson2017clevr} in \ac{acre}'s design and add a Blicket machine to signal its state of activation, intentionally simplifying visual information processing and emphasizing causal reasoning. Following attempts made in abstract spatial-temporal reasoning benchmarks~\cite{hu2020hierarchical,santoro2018measuring,zhang2019raven}, we provide the visual reasoning system with sets of panel images as context and use image-based queries to ease language understanding, echoing the setup and the learning theories in developmental literature~\cite{gopnik2012scientific,gopnik2004theory,gopnik2000detecting,gopnik2001causal,gopnik2012reconstructing}.

Specifically, each problem in \ac{acre} consists of $10$ panels: $6$ for context and $4$ for query. The $6$ context panels are divided into two sets, the first of which serves as an introduction to the Blicket mechanism that some objects activate the machine, and others do not. This simpler set of panels resembles the introductory trials administered to children in human experiments~\cite{gopnik2001causal,sobel2004children}. Instead of bringing in the concept of Blicket\footnote{While the notion of ``Blicket'' is not necessary for a visual reasoning system to solve the task, we use the term throughout this paper to simplify expressions and facilitate understanding of the core ideas.}, in queries, we only ask a visual reasoning system to predict the state of the Blicket machine given the objects in the queries. Half of the queries concern the independent scenarios, wherein a single object is presented, and the system is challenged to reason about whether this object is one of the causes that could activate the Blicket machine. The remaining half of the queries are for interventional scenarios, wherein we intervene in an existing context panel and ask what the state of the Blicket machine would be under the intervention. Each query is independent such that statistical bias~\cite{gopnik2001causal,sobel2004children} and potential cheating for abstract reasoning~\cite{hu2020hierarchical,zhang2019raven} are minimized. In summary, \ac{acre} includes $30,000$ abstract causal reasoning problems, supports all $4$ types of reasoning queries (direct, indirect, screening-off, and backward-blocking), and is fully annotated with object attributes, bounding boxes, and masks. We further design two \ac{ood} generalization splits in \ac{acre} to evaluate models' generalizability.

In experiments, we use the \ac{acre} dataset to analyze current visual reasoning systems' ability in causal induction. Despite remarkable results in other visual reasoning tasks, we notice that pure neural networks~\cite{devlin2019bert,he2016deep,santoro2018measuring,wang2020abstract,zheng2019abstract} favor a covariation-based reasoning strategy and thus can only achieve performance marginally above the chance level. As the first attempt in the exploration to empower visual reasoning systems for causal induction, we resort to neuro-symbolic models~\cite{han2019visual,li2020closed,mao2019neuro,qi2020generalized,qi2018generalized,yi2020clevrer,yi2018neural,zhang2021abstract,zhang2020machine} that combine neural visual processing~\cite{he2017mask} and symbolic causal reasoning~\cite{glymour2019review,pearl1995theory,rescorla1972theory,spirtes2000causation,zheng2018dags,zheng2020learning}, which turn out to struggle in backward-blocking cases in abstract causal reasoning. 

To sum up, this paper makes three primary contributions:
\begin{itemize}[leftmargin=*,noitemsep,nolistsep]
    \item We propose the \acf{acre} dataset to probe current visual reasoning systems' capacity in causal induction. The dataset is inspired by the Blicket experiments and contains $30,000$ problems. \ac{acre} covers all $4$ types of causal reasoning queries (direct, indirect, screening-off, and backward-blocking) with additional \ac{ood} generalization splits.
    \item We benchmark and analyze state-of-the-art visual reasoning models in \ac{acre}. Experimental results show that neural models tend to capture statistical correlations in observation but fail to induce the underlying causal relationships demonstrated in the trials.
    \item We propose neuro-symbolic combinations that improve on pure neural networks. However, our analysis shows that even with the inductive bias in causality, they still fail to distinguish a true cause from superficial covariation in backward-blocking cases. Taken together, these deficiencies call for future research in models with a more comprehensive capability of causal induction.
\end{itemize}

\begin{figure*}[t!]
    \centering
    \includegraphics[width=\linewidth]{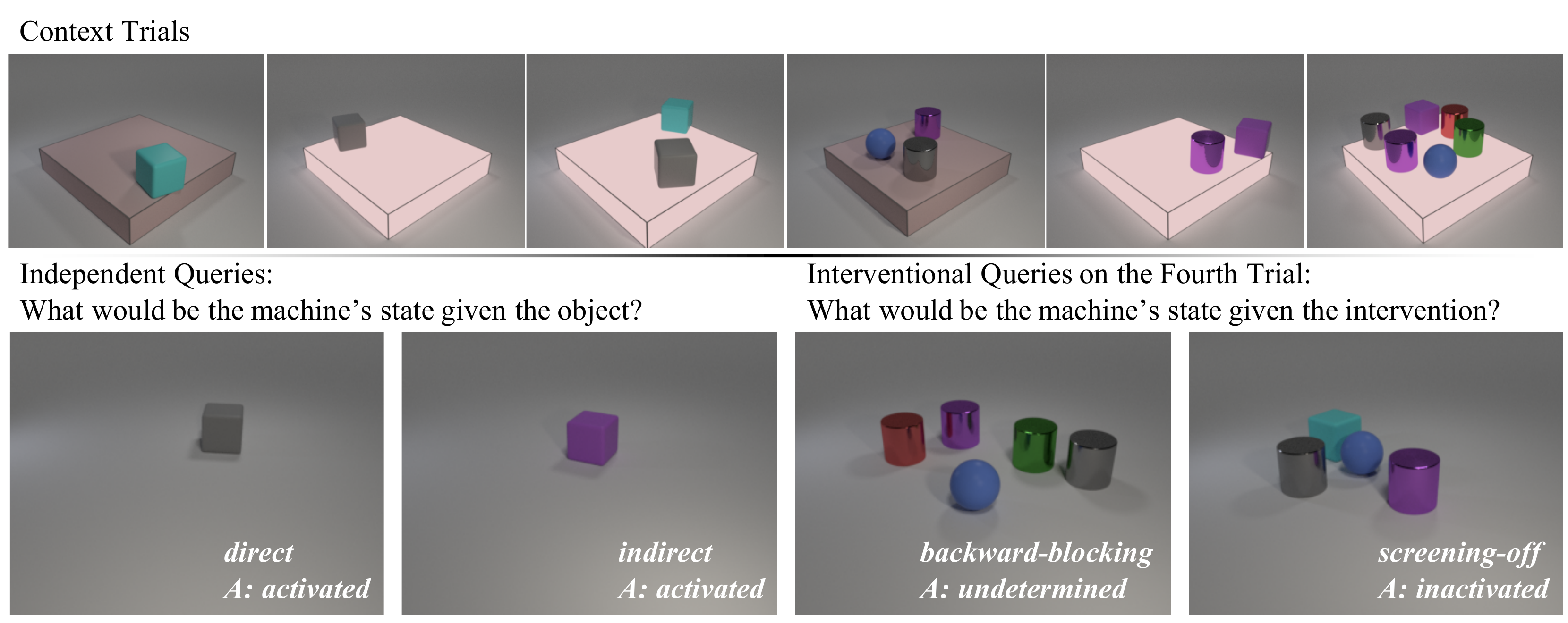}
    \caption{A sample problem in \ac{acre}. Of the $6$ context trials, we devote the first set of $3$ panels for an introduction to the Blicket machinery and allow more complex configurations in the second set of panels. Queries are either on independent objects or interventional combinations for an existing trial. In this example, the first query tests causal reasoning from \textit{direct} evidence, as the gray cube is independently tested and always associated with an activated machine. The second query requires comparing the fourth and fifth trial to realize that the Blicket machine is activated by the cube, not the cylinder, based on \textit{indirect} evidence. As such, we infer that the red and green cylinders in the sixth trial may not activate the machine because the purple cube can already do so; despite their association with an activated machine only, their Blicketness is \textit{backward-blocked} in the interventional trial. The cyan cube is \textit{screened-off} by the gray cube's Blicketness from probabilistically activating the machine. Of note, the screening-off and the backward-blocking case cannot be solved by covariation.}
    \label{fig:example}
\end{figure*}

\section{Related Work}

\paragraph{Abstract Visual Reasoning}

To date, the computer vision and \ac{ai} community's efforts in abstract visual reasoning primarily focus on the specific task of \ac{rpm}~\cite{carpenter1990one,raven1938raven}, commonly known as \ac{iq} tests, that studies how visual reasoning systems can induce the hidden spatial-temporal transformation from limited context and apply it to derive a missing panel. Santoro \etal~\cite{santoro2018measuring} extended the relational module~\cite{santoro2017simple} to take panel-based representation and introduced the \ac{wren}. Zhang \etal~\cite{zhang2019raven} proposed to incorporate structural annotations in a neural modular manner. Methods considering contrast at data-level~\cite{hill2018learning} or module-level~\cite{zhang2019learning} were later shown to improve performance significantly. Zheng \etal~\cite{zheng2019abstract} formulated the problem as teacher-student learning, Wang \etal~\cite{wang2020abstract} used a multiplex graph model to capture the hidden relations, and Spratley \etal~\cite{spratley2020closer} revisited ResNet models combined with unsupervised learning. More recently, Zhang \etal~\cite{zhang2021abstract} disentangled perception and reasoning from a monolithic model, wherein the visual perception frontend predicts objects' attributes, later aggregated by a scene inference engine to produce a probabilistic scene representation, and the symbolic logical reasoning backend abduces the hidden rules.

The proposed \ac{acre} dataset complements the spectrum of abstract visual reasoning tasks by challenging visual reasoning systems with causal induction from a limited number of trials and adding missing dimensions of causal understanding into the prior spatial-temporal task set.

\paragraph{Causal Reasoning}

Equipping visual reasoning systems with causal reasoning capability has been an emerging topic in computer vision research~\cite{fire2015learning,lopez2017discovering,mottaghi2016happens}. Recent causal reasoning datasets~\cite{baradel2020cophy,yi2020clevrer} established video-based benchmarks\footnote{Of note, these prior works do not echo Michotte's theory of perceived causality that humans possess a ``causal detector'' akin to how we perceive colors~\cite{michotte2017perception}, as they fail to show humanlike causal perception~\cite{scholl2000perceptual,zhu2020dark}.} for either trajectory prediction in counterfactual scenarios or visual question answering with explanatory, predictive, and counterfactual questions. Nevertheless, causal induction in prior computer vision research relies heavily on covariation. For instance, psychological research~\cite{battaglia2013simulation,bramley2018intuitive,gerstenberg2017eye,gerstenberg2017intuitive,kubricht2017intuitive} points out that the key to solving these two problems is intuitive physics, with covariation-based causal reasoning in associating collision with object dynamics. Moreover, Edmonds \etal~\cite{edmonds2019decomposing,edmonds2018human,edmonds2020theory} further demonstrate that covariation would result in catastrophic failures when the visual features are similar but the underlying causal mechanisms dramatically differ. These results necessitate causal induction beyond covariation: Asymmetries in learning under various causal structures~\cite{waldmann1992predictive} refute parsimonious associative learning~\cite{shanks1988associative}.

With a particular emphasis on causal induction beyond the simple causal reasoning strategy of covariation~\cite{lagnado2007beyond}, we design \ac{acre} with diversified causal queries, requiring a visual reasoning system to induce the hidden causal mechanism from only limited observation. From a cognitive standpoint, it is argued that Bayesian networks~\cite{pearl2009causality,pearl1995theory} and the theory-theory~\cite{edmonds2020theory,gopnik2012scientific,gopnik2004theory,gopnik2012reconstructing,griffiths2009theory,sobel2006blickets} play vital roles in abstract causal reasoning. However, how young toddlers induce accurate Bayesian representation and form a correct theory during such a short exposure remains unclear~\cite{frosch2012causal}.

\section{Building \acs{acre}}\label{sec:acre}

The \ac{acre} dataset is designed to be light in visual recognition and heavy in causal induction. Specifically, we ground every panel in a fully-controlled synthetic environment by adopting the CLEVR universe~\cite{johnson2017clevr} where all objects, including the Blicket machine, are placed on a tabletop with three-point lighting. All potential Blicket objects are of the same size and come with $3$ possible shapes (cube, sphere, or cylinder), $2$ possible materials (metal or rubber), and $8$ possible colors (gray, red, blue, green, brown, cyan, purple, or yellow). For the context panels, we set all objects on a pink Blicket machine at the center of the scene and signal its state of activation by lighting it up. For the query panels, we directly put all objects on the tabletop. In both cases, objects are randomly distributed across the scenes. To avoid confusion during reference, every object is uniquely identifiable by its shape, material, and color. Other than the constraints, every object's attributes are randomly sampled from the aforementioned space. Collectively, every \ac{acre} problem contains $5$ to $8$ unique objects. We keep other scene configurations the same as the original setup in CLEVR~\cite{johnson2017clevr} and generate images by Blender~\cite{blender2016blender}. Every image is also fully-annotated with object attributes, bounding boxes, and masks; see \cref{fig:example} for a sample problem in \ac{acre} and refer to the supplementary for more examples.

\paragraph{\ac{acre} Context}

Every \ac{acre} problem contains $10$ panels, of which $6$ serve as context panels. Following the original design~\cite{gopnik2001causal,sobel2004children}, we further divide the $6$ panels into $2$ sets and use the first simpler set as the familiarization set. Specifically, for the first set of $3$ panels, we randomly sample $2$ objects and assign one to be a Blicket and another to be a non-Blicket. Both objects are independently tested on the Blicket machine and then placed together on it. These $3$ simple trials reveal the nature of a Blicket detector: The machine will be activated when a Blicket is placed on it. For the second set of panels, we allow more random sampling; in particular, we sample another group of objects that is disjoint with the first one and partition it into $3$ potentially overlapping subgroups, corresponding to the configurations for each of the rest panels. Either one or two of them are associated with an activated Blicket machine.

\begin{figure}[t!]
    \centering
    \begin{overpic}[width=\linewidth]{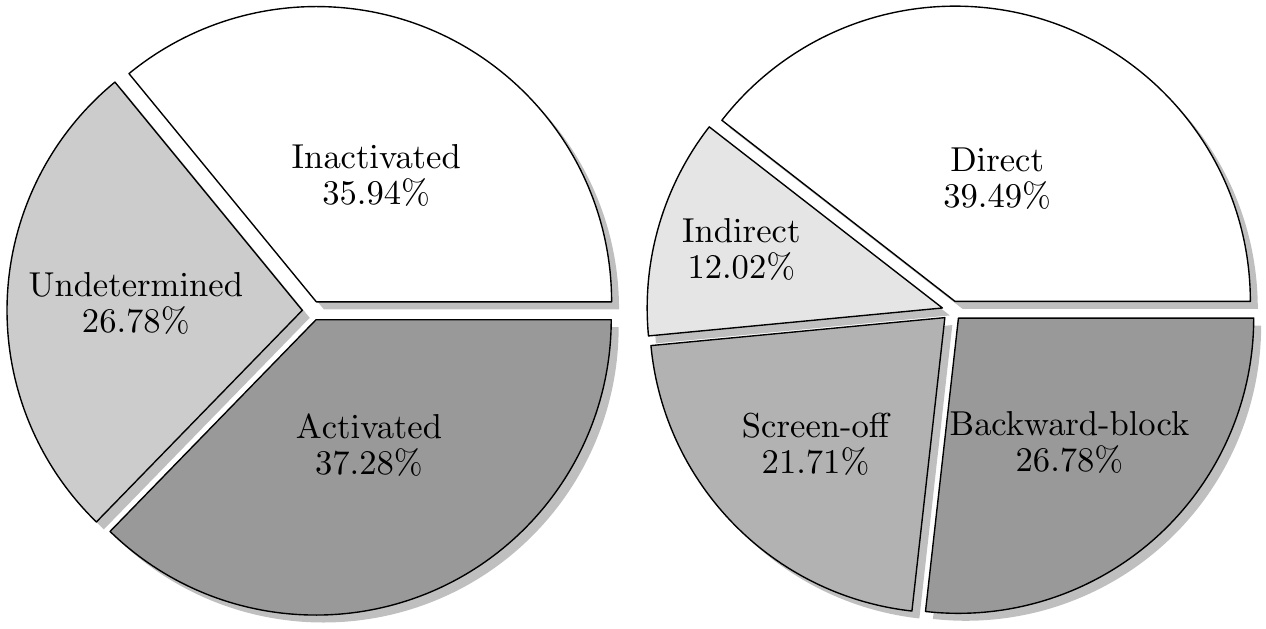}
        \put(1,45){(a)}
        \put(50,45){(b)}
    \end{overpic}
    \caption{Distributions of (a) labels and (b) query types in \ac{acre}.}
    \label{fig:stats}
\end{figure}

\paragraph{\ac{acre} Query}

The Blicket machine's activation pattern in context panels supports all $4$ types of queries (direct, indirect, screening-off, and backward-blocking) and provide sufficient clues for determining Blicketness for each object; see illustrations in \cref{fig:intro,fig:example}. Based on explanations of the Blicket mechanism~\cite{gopnik2004theory,gopnik2001causal,gopnik2012reconstructing,sobel2004children}, we detail query categorization in the following. Intuitively, an object is a Blicket if it is independently and always associated with an activated machine, which can be determined based on direct evidence; the same reasoning strategy is applicable to resolve non-Blickets. An object is also considered a Blicket if the machine activates when we place it together with other objects but not alone, and the other objects fail to activate the machine. In these cases, the Blicketness is resolved by indirect evidence; no direct observation is available. An object is considered a non-Blicket when putting it with other potential Blickets together will activate the machine, but it fails to do so by itself; this derivation is referred to as screening-off reasoning. In addition to being a Blicket or non-Blicket, an object's Blicketness could also be undetermined, which occurs when the object is not directly tested, but can activate the machine together with other potential Blickets; this is referred to as backward-blocking reasoning. Note that the Blicketness of an individual object may be undetermined, but together with other undetermined ones, they can form a set that activates the machine; as such queries happen in the indirect setting, we also refer to them as indirect reasoning.

The rich causal relations embedded in the context panels afford us to probe a reasoning system's causal induction capability. In particular, we design $4$ queries in each \ac{acre} problem, $2$ for independent scenarios and another $2$ for interventional scenarios, similar to the questions administered in human experiments~\cite{gopnik2001causal,sobel2004children}. In the independent scenario, we randomly sample one object from those tested in the trials. In the interventional scenario, we pick a trial with an inactivated machine and add a set of objects randomly picked from those in the context panels. The reasoning system is then asked to tell the status of the Blicket machine after placing the objects on it, either inactivated, undetermined, or activated. To avoid statistical bias~\cite{gopnik2001causal,sobel2004children} or potential cheating~\cite{hu2020hierarchical,zhang2019raven}, all queries in a problem are independent.

\paragraph{Generalization Splits}

\ac{acre} comes with additional \ac{ood} splits to measure model generalization in causal induction; we focus on compositionality and systematicity in systematic generalization~\cite{fodor1988connectionism,gordon2019permutation,lake2017building,xie2021halma}. In the compositionality split, we assign different shape-material-color combinations to the training and test set and ensure the training set contains every shape, material, and color, similar to the Compositional Generalization Test (CoGenT) in CLEVR~\cite{johnson2017clevr}. In the systematicity split, we vary the distribution of an activated Blicket detector in the context panels, with the machine lighting up $3$ times in the training set and $4$ times during testing. Note the strategy for causal induction remains the same irrespective of the distribution change.

\begin{figure*}[t!]
    \centering
    \includegraphics[width=\linewidth]{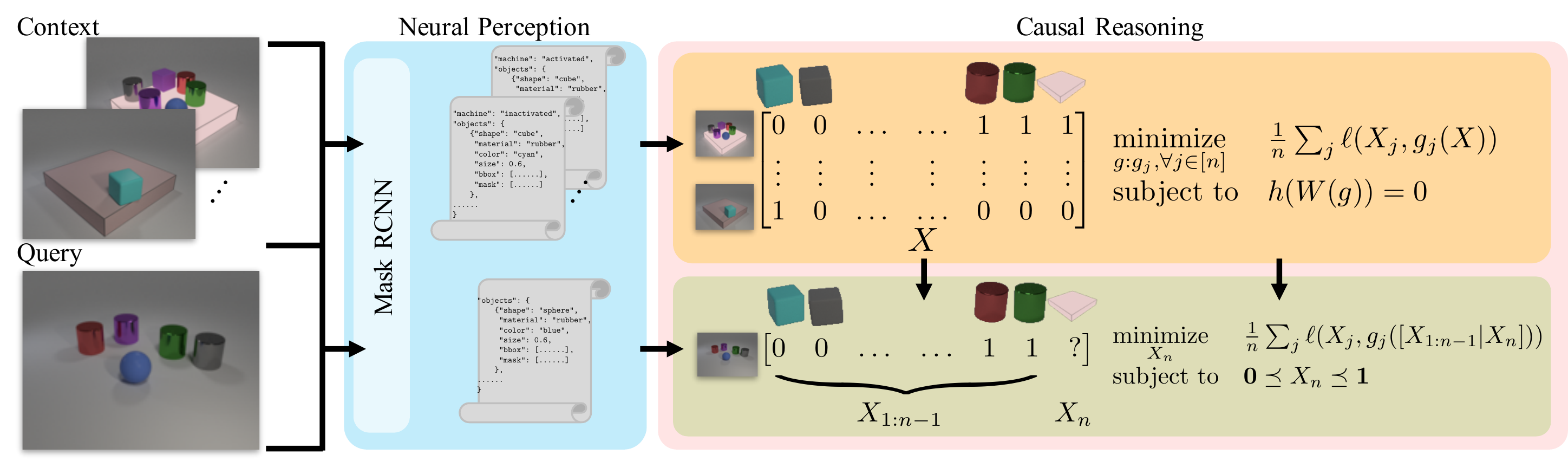}
    \caption{An illustration of the proposed neuro-symbolic combination (NS-Opt) for \ac{acre}. The neural frontend is responsible for scene parsing. In particular, we use a Mask RCNN to detect objects and classify their attributes as well as the Blicket machine's state. The parsed results are arranged into data matrices and sent into the causal reasoning backend for optimization. A generalized \ac{sem} is learned from context trials during reasoning, which is further used to infer the state of the Blicket machine for each query.}
    \label{fig:model}
\end{figure*}

In total, \ac{acre} contains $30,000$ problems, evenly partitioned into an \ac{iid} split, a compositionality split, and a systematicity split. The dataset covers all $4$ types of queries, and the label distribution is adjusted to be roughly uniform; see \cref{fig:stats} for the label distribution and query type distribution in the dataset. Please refer to the supplementary for detailed distributions of labels and query types for each split.

\section{Reasoning Systems on \acs{acre}}

This section details the deep neural models adopted to benchmark the proposed \acs{acre} dataset and the neuro-symbolic combinations explicitly designed to incorporate inductive bias for causal induction.

\subsection{Deep Neural Models}

As \ac{acre} shares the inductive nature with \ac{rpm}, we test several established models designed for it~\cite{santoro2018measuring,wang2020abstract,zheng2019abstract}. We also test methods commonly used for linguistic or visual modeling~\cite{devlin2019bert,he2016deep}. Each context-query pair is independently fed into the network and treated as a classification problem.

\paragraph{CNN-\acs{mlp}} We concatenate context panels with the query panel in the channel dimension and use a $4$-layer standard CNN architecture to extract features. The CNN architecture interleaves batch normalization and ReLU activation between convolution layers. The final convolved features are passed to a $2$-layer \acf{mlp} with a dropout layer with a rate of $0.5$ in between the two layers.

\paragraph{ResNet-\acs{mlp}} In this model, we replace the CNN backbone in CNN-\ac{mlp} with ResNet-18~\cite{he2016deep}.

\paragraph{CNN-LSTM} We use a standard LSTM model~\cite{hochreiter1997long} to further process visual features. Specifically, we independently extract image features of each panel using a CNN, append a one-hot position tag to each feature map, and pass them sequentially into the LSTM module. The final hidden state is further processed by a linear layer to produce logits. 

\paragraph{CNN-BERT} A visual BERT~\cite{devlin2019bert} model is also tested. We compute image features using a CNN and follow practices in BERT: For the sequence of image features, we prepend \texttt{<CLS>}, separate the context panels and the query panel with \texttt{<SEP>}, and add position and segment embeddings. Output from \texttt{<CLS>} is then used for classification.

\paragraph{\ac{wren}} We adopt the \ac{wren} model proposed by Santoro \etal~\cite{santoro2018measuring}, which applies the relational module~\cite{santoro2017simple} on panel-based image representations.

\paragraph{LEN} LEN~\cite{zheng2019abstract} stems from \ac{wren} but takes into account the row-wise and column-wise compositions in \ac{rpm}, deeper features, and the multi-choice setting. We adapt the original LEN design to the proposed \ac{acre} by removing branches for the column-wise composition and making the prediction on each query independent.

\paragraph{MXGNet} A similar strategy in LEN is used to make MXGNet~\cite{wang2020abstract} compatible with \ac{acre}. The two sets of context trials are treated as rows in the model.

\subsection{Neuro-Symbolic Models}

In preliminary experiments, we notice that pure neural models tend to capture statistical correlation rather than modeling the hidden causal relations out of the context trials. To overcome this issue, we propose neuro-symbolic combinations and explicitly incorporate various forms of causal inductive bias for the abstract causal reasoning task.

Specifically, we draw inspirations from recent advances in neuro-symbolic literature~\cite{han2019visual,li2020closed,mao2019neuro,qi2020generalized,qi2018generalized,yi2020clevrer,yi2018neural,zhang2021abstract,zhang2020machine} and decompose our model into a neural perception frontend and a causal reasoning backend. By design, the frontend is responsible for parsing each context trial to form an object-based representation, whereas the backend takes the symbolic output from the frontend and performs causal induction; see an overview of the method in \cref{fig:model}.

\paragraph{Neural Perception Frontend} As the first attempt to solve \ac{acre} problems, we disentangle our neural perception frontend and independently pretrain the model to parse each scene. Specifically, we use Mask-RCNN~\cite{he2017mask} with ResNet-50 FPN~\cite{he2016deep,lin2017feature} backbone. The perception model is tasked with predicting the Blicket machine's state, object masks, and object attributes (shape, material, and color) for each object in the scene. Both context and query panels in the training set of each split are used to train the frontend.

\paragraph{Causal Reasoning Backend} Due to its efficiency and accuracy, we use a score-based continuous optimization method, denoted as \textbf{NS-Opt}, to simultaneously learn a generalized \acf{sem} and derive the hidden causal relations~\cite{zheng2018dags,zheng2020learning}. In particular, denoting the existence of object $j$ in panel $i$ as $X_{i, j} \in \{0, 1\}$, we can arrange the symbolic parsing results from the neural perception frontend into a data matrix $X \in \{0, 1\}^{6 \times n}$, where $n$ equals the number of unique objects in all context panels plus the Blicket machine. A generalized \ac{sem} assumes that the state of object $j$ is related to states of its parents via a function and can be represented as
\begin{equation}
    X_{j} = f_j(X_{\text{pa}(j)}) = g_j(X),
\end{equation}
where $X = [X_1 | X_2 | \ldots | X_j | \ldots]$, and $\text{pa}(j)$ denotes the parents of object $j$. The parent finding process is further generalized in $g_j(\cdot)$ and put into optimization constraints. 

Following~\cite{zheng2020learning}, we formulate causal discovery as an optimization problem
\begin{align}
    \begin{array}{l l}
        \underset{g: g_j, \forall j \in [n]}{\text{minimize}} & \frac{1}{n} \sum_j \ell(X_j, g_j(X)) \\
        \text{subject to}                                     & h(W(g)) = 0,
    \end{array}
\end{align}
where $W(g)_{k, j} = \norm{\partial_k g_j} \forall k, j \in [n]$, and $h(W) = \text{Tr}(e^{W \circ W} - I)$. We use $[n]$ to denote an integer set from $1$ to $n$, $\norm{\cdot}$ the $L^2$ function norm, and $\circ$ the Hadamart product. Using the binary cross entropy loss as $\ell(\cdot, \cdot)$ for each object $j$, the optimization problem regularizes the generalized \ac{sem} to reconstruct the observation, while constraining the relations among the variables to be a causal \ac{dag}: $W(g)$ can be regarded as the adjacency matrix among variables, and $h(\cdot)$ a metric for acyclicity. We use an \ac{mlp} for each $g_j(\cdot)$ and optimize the problem by Augmented Lagrangian; see~\cite{zheng2018dags,zheng2020learning} for details.

With a learned generalized \ac{sem} representing the hidden causal relations in the context trials, we treat each query as another optimization problem. Specifically, we construct a partial data vector for each panel from the symbolic representation parsed by the neural perception frontend. Denoting the Blicket machine as object $n$, the query vector can be represented as $X_{1:n - 1}$. Treating $X_n$ as the probability of the Blicket machine being activated, the query optimization is formulated as 
\begin{align}
    \begin{array}{l l}
        \underset{X_n}{\text{minimize}} & \frac{1}{n} \sum_j \ell(X_j, g_j([X_{1:n - 1} | X_n])) \\
        \text{subject to}               & \mathbf{0} \preceq X_n \preceq \mathbf{1}.
    \end{array}
\end{align}
We solve it using L-BFGS-B~\cite{byrd1995limited} and set thresholds on $X_n$ to predict the final state of the Blicket machine.

We also test a constraint-based method~\cite{glymour2019review,spirtes2000causation} and the well-known \ac{rw} model~\cite{rescorla1972theory} for the causal reasoning backend. The constraint-based method (denoted as \textbf{NS-PC}) first uses the state-of-the-art PC algorithm~\cite{spirtes2000causation} to test conditional independence and search for an underlying causal \ac{dag} among objects and the Blicket machine. It then finds the parent nodes for the Blicket machine and estimates its conditional probability table, which can be directly read out for each query configuration. For the \ac{rw} model (denoted as \textbf{NS-RW}), we simply treat co-occurrence of an object with an activated Blicket machine as its Blicketness. A query configuration's state is predicted based on the maximum Blicketness of all objects in it.

\section{Experiments}

\subsection{Experimental Setup}

\ac{acre} is equally partitioned into $3$ splits, \ie, the \ac{iid} split, the compositionality (comp) split, and the systematicity (sys) split. Each of the splits contains $10,000$ problems. We further divide each split into $10$ folds, with $6$ folds for training, $2$ folds for validation, and $2$ folds for testing. All models are trained on the training sets, with hyper-parameters tuned on the validation sets. Results are reported for the best models on the test sets. In particular, we report two metrics: query accuracy and problem accuracy. The former measures how a model performs on \textit{each} query, and the later whether a model correctly answers \textit{all} 4 queries in a problem instance. Note that based on the label distribution shown in \cref{fig:stats}, a simple strategy of always predicting activation will yield around $37.3\%$ query accuracy and $1.87\%$ problem accuracy, and a completely random guess would yield $33.3\%$ query accuracy and $1.19\%$ problem accuracy. All neural models, including the neural perception frontend in the neuro-symbolic models, are implemented in PyTorch~\cite{paszke2017automatic} and optimized using Adam~\cite{kingma2014adam}. All experiments were run on an Nvidia Titan XP GPU.

\subsection{Performance on the \ac{iid} Setting}

\begin{table*}[t]
    \centering
    \resizebox{\linewidth}{!}{
        \begin{tabular}{ c | c | c  c  c  c  c  c  c | c  c  c}
            \hline
            \multicolumn{2}{c|}{Method}        & MXGNet    & LEN       & CNN-\ac{mlp}   & WReN      & CNN-LSTM  & ResNet-\ac{mlp} & CNN-BERT  & NS-RW     & NS-PC     & \textbf{NS-Opt}             \\
            \hline\hline
            \multirow{2}{*}{\ac{iid}}   & Qry. & $33.01\%$ & $38.08\%$ & $40.86\%$ & $40.39\%$ & $41.91\%$ & $42.00\%$  & $43.56\%$ & $46.61\%$ & $59.26\%$ & $\mathbf{66.29\%}$ \\
                                        & Pro. & $1.00\%$  & $2.05\%$  & $3.25\%$  & $2.30\%$  & $3.60\%$  & $3.35\%$   & $3.50\%$  & $6.45\%$  & $21.15\%$ & $\mathbf{27.00\%}$ \\
            \hline
            \multirow{2}{*}{Comp.}      & Qry. & $35.56\%$ & $38.45\%$ & $41.97\%$ & $41.90\%$ & $42.80\%$ & $42.80\%$  & $43.79\%$ & $50.69\%$ & $61.83\%$ & $\mathbf{69.04\%}$ \\
                                        & Pro. & $1.55\%$  & $2.10\%$  & $2.90\%$  & $2.65\%$  & $2.80\%$  & $2.60\%$   & $2.40\%$  & $8.10\%$  & $22.00\%$ & $\mathbf{31.20\%}$ \\
            \hline
            \multirow{2}{*}{Sys.}       & Qry. & $33.43\%$ & $36.11\%$ & $37.45\%$ & $39.60\%$ & $37.19\%$ & $37.71\%$  & $39.93\%$ & $42.18\%$ & $62.63\%$ & $\mathbf{67.44\%}$ \\
                                        & Pro. & $0.60\%$  & $1.90\%$  & $2.55\%$  & $1.90\%$  & $1.85\%$  & $1.75\%$   & $1.90\%$  & $4.00\%$  & $29.20\%$ & $\mathbf{29.55\%}$ \\
            \hline
        \end{tabular}%
    }%
    \caption{Performances of models on the \ac{iid} split, the compositionality split (Comp.), and the systematicity split (Sys.) in \ac{acre}. We report $2$ evaluation metrics: query accuracy (Qry.) and problem accuracy (Pro.). Please refer to the experimental setup for details.}
    \label{tab:perf}
\end{table*}

The first portion of \cref{tab:perf} reports how various models perform under the \ac{iid} setting of \ac{acre}. Surprisingly, existing state-of-the-art methods for the abstract spatial-temporal reasoning task~\cite{wang2020abstract,zheng2019abstract} do not fare much better (even worse in certain cases) than a simple CNN-\ac{mlp} model. In particular, MXGNet performs slightly worse than a random guess, only correctly answering $1\%$ of problems. With a relational module, \ac{wren} is on par with the CNN-\ac{mlp} model. CNN-LSTM and ResNet-\ac{mlp} achieve similar performance, with the LSTM-based reasoning model performing better in problem accuracy. Of all pure neural models, the BERT model achieves the best in query accuracy, slightly overtaken by CNN-LSTM in problem accuracy.

Among the $3$ neuro-symbolic models tested, NS-RW strictly follows the covariation strategy in solving the causal discovery problems. We notice that such a simple causal reasoning method can only handle less than half of \ac{acre} queries and less than $10\%$ of \ac{acre} problems, verifying and necessitating our efforts to create a benchmark for causal induction beyond covariation. NS-PC serves as an oracle model for causal discovery as it adopts independence tests and search methods. However, our experiments show that NS-PC is inferior to the optimization-based NS-Opt method. We believe such a result is due to the sparse and limited observation in \ac{acre} problems, making it difficult to perform reliable independence tests. This challenge further perplexes the underlying mechanisms on how humans, even toddlers, derive the hidden relations so quickly and accurately from scarce observation. The proposed NS-Opt method successfully handles two-thirds of the queries but still has much room to improve on problem accuracy. 

\subsection{Performance on the \ac{ood} Settings}

The second and third portions of \cref{tab:perf} depict the models' performance on the \ac{ood} settings, \ie, compositionality and systematicity. Comparing both query accuracy and problem accuracy in the compositionality split with those in the \ac{iid} setting, we notice that models' performances have no significant changes. Considering the fact that the training set and the test set in the compositionality split contain completely different object attribute combinations, it is likely that neural models indeed have emerged a certain level of causal reasoning, though not perfect, to solve the problems, rather than entirely relying on statistical visual features from the training set. However, their underlying representation of causal knowledge is still elusive; future work is in need to discover their precise mechanisms.

Even if neural models emerged a causal reasoning strategy, such a strategy is not systematic, as demonstrated in the comparison between the systematicity split and the \ac{iid} split. Note that only the distributions of an activated machine are different in the training set and the test set of the systematicity split, while the solutions can be derived in the same way. We note that except for the NS-PC model and the NS-Opt model, all other models experience performance drop; some of them are even worse than always predicting ``activated''. This observation echoes the recent empirical results that pure neural models still struggle to systematically generalize~\cite{fodor1988connectionism,lake2018generalization,ruis2020benchmark}.

\begin{table*}[t]
    \centering
    \resizebox{\linewidth}{!}{
    \begin{tabular}{c | c | c  c  c  c  c  c  c | c  c  c}
        \hline
        \multicolumn{2}{c|}{Method}        & MXGNet    & LEN       & CNN-\ac{mlp}   & WReN      & CNN-LSTM  & ResNet-\ac{mlp} & CNN-BERT  & NS-RW     & NS-PC     & \textbf{NS-Opt}    \\
        \hline\hline
        \multirow{4}{*}{\ac{iid}}   & D.R. & $27.73\%$ & $49.07\%$ & $55.56\%$ & $51.04\%$ & $48.20\%$ & $54.87\%$  & $52.24\%$ & $88.88\%$ & $84.46\%$ & $\mathbf{91.64\%}$ \\
                                    & I.D. & $29.63\%$ & $45.11\%$ & $56.31\%$ & $41.04\%$ & $36.76\%$ & $48.37\%$  & $44.50\%$ & $\mathbf{99.29\%}$ & $29.33\%$ & $69.25\%$ \\
                                    & S.O. & $14.88\%$ & $33.68\%$ & $44.88\%$ & $29.75\%$ & $53.23\%$ & $42.29\%$  & $42.59\%$ & $7.21\%$  & $78.31\%$ & $\mathbf{85.37\%}$ \\
                                    & B.B. & $\mathbf{59.09\%}$ & $23.91\%$ & $9.71\%$  & $35.61\%$ & $24.91\%$ & $21.12\%$  & $32.15\%$ & $1.66\%$  & $20.50\%$ & $11.98\%$ \\
        \hline
        \multirow{4}{*}{Comp.}      & D.R. & $36.93\%$ & $47.58\%$ & $57.59\%$ & $55.29\%$ & $56.58\%$ & $62.79\%$  & $54.07\%$ & $91.74\%$ & $89.50\%$ & $\mathbf{92.50\%}$ \\
                                    & I.D. & $55.99\%$ & $52.51\%$ & $64.38\%$ & $66.94\%$ & $65.10\%$ & $70.01\%$  & $46.88\%$ & $\mathbf{99.80\%}$ & $28.66\%$ & $76.05\%$\\
                                    & S.O. & $0.00\%$  & $18.01\%$ & $31.66\%$ & $8.44\%$  & $19.69\%$ & $30.52\%$  & $40.57\%$ & $4.07\%$  & $85.28\%$ & $\mathbf{88.33\%}$\\
                                    & B.B. & $\mathbf{52.35\%}$ & $33.63\%$ & $15.26\%$ & $35.99\%$ & $29.27\%$ & $8.54\%$   & $28.79\%$ & $0.67\%$  & $15.21\%$ & $13.48\%$\\
        \hline
        \multirow{4}{*}{Sys.}       & D.R. & $15.24\%$ & $46.22\%$ & $70.79\%$ & $53.56\%$ & $42.57\%$ & $65.19\%$  & $55.97\%$ & $92.44\%$ & $89.76\%$ & $\mathbf{94.73\%}$ \\
                                    & I.D. & $5.42\%$  & $47.90\%$ & $87.61\%$ & $71.35\%$ & $37.61\%$ & $85.07\%$  & $68.25\%$ & $\mathbf{99.89\%}$ & $57.08\%$ & $88.38\%$ \\
                                    & S.O. & $42.58\%$ & $30.91\%$ & $11.57\%$ & $16.80\%$ & $63.28\%$ & $9.57\%$   & $0.00\%$  & $0.20\%$  & $73.93\%$ & $\mathbf{82.76\%}$ \\
                                    & B.B. & $\mathbf{56.38\%}$ & $24.89\%$ & $3.60\%$  & $31.62\%$ & $8.70\%$  & $13.38\%$  & $45.59\%$ & $0.46\%$  & $24.88\%$ & $16.06\%$ \\
        \hline
    \end{tabular}
    }
    \caption{A closer look at how models perform on each type of queries on different splits of \ac{acre}: direct (D.R.), indirect (I.D.), screening-off (S.O.), and backward-blocking (B.B.).}
    \label{tab:q_type}
\end{table*}

Across the $3$ splits, we also notice the conspicuously large gap between the query accuracy and problem accuracy. We hypothesize that the result indicates the existence of the bucket effect, which we verify in the next section.

\subsection{A Closer Look at Queries}

\setstretch{0.98}

The drastic difference between query accuracy and problem accuracy motivates us to perform a closer inspection on how models perform on each type of queries; see \cref{tab:q_type} for a summary of our experimental results.

In general, we notice that neural models tend to capture causal relations by covariation. Most of them excel in query types directly solvable by this strategy, achieving the best performance in direct queries or indirect queries or both across the different splits. This effect is particularly significant in CNN-based reasoning models (CNN-\ac{mlp} and ResNet-\ac{mlp}) that even reach $87\%$ accuracy for indirect queries by learning from only target labels. However, in contrast to the satisfactory performance on covariation-based reasoning, they are unable to handle the screening-off queries and the backward-blocking queries, which go beyond co-occurrence. Specifically, the best-performing neural model (CNN-BERT) embarrassingly fails on screening-off queries in the systematicity split, while CNN-based reasoning models also struggle in these settings. Among the relation-module-based models (MXGNet, LEN, and \ac{wren}), LEN and \ac{wren} are relatively stable across the different types of queries. However, with a multiplex graph, MXGNet shows different dynamics, learning best in the backward-blocking queries but counter-intuitively underperforming in the direct and indirect queries. It is also worth noting that causal reasoning that supports backward-blocking for MXGNet does not consistently enable screening-off reasoning. A converse observation is found in CNN-LSTM: The model shines in screening-off reasoning but fades in backward-blocking in $2$ of the splits. Taking these results together, we hypothesize that pure neural visual reasoning systems have not yet mastered causal induction to a comparable level humans displayed in the developmental studies~\cite{gopnik2001causal,sobel2004children}.

Performance differences in queries among the neuro-symbolic models potentially point out an Achilles' heel for solving abstract causal reasoning problems. NS-RW's inferior performance is expected as the model only considers covariation and will surely fail the screening-off and backward-blocking queries, despite its success in direct and indirect queries. NS-RW's results also serve as a sanity check for queries in \ac{acre} that nearly all of the direct and indirect queries can be solved by covariation (except for a minimum number of interventional cases) and nearly none of the screening-off and backward-blocking queries can (except for a minimum number of coincidences). Comparing NS-PC and NS-Opt, we notice that both models achieve fair performance on direct queries and the screening-off queries. However, the latter fares significantly better in indirect queries. We argue that the strict independence tests and search methods used in PC make the model less robust against noise, especially under the sparse-and-limited-data scenario. What is evident in both models, and more significant in NS-Opt, is their inability in differentiating the superficial correlation with an activated machine and the undetermined Blicketness within. This close inspection also indicates that adequately addressing the issue can further improve the general causal reasoning performance. By comparing the low accuracy of NS-Opt and pure neural networks in backward-blocking, we hypothesize that a potential solution to causal reasoning would be to combine the best of both worlds in learning and symbolic reasoning, keeping both the learnability of neural methods and the interpretability of symbolic methods.

\section{Conclusion}

In this work, we present a new dataset for \acf{acre}, aiming to measure and improve causal induction in visual reasoning systems. Apart from the inductive reasoning nature, the defining feature of the \ac{acre} dataset is the requirement to perform causal reasoning beyond covariation. Inspired by the established stream of research on Blicket experiments, the \ac{acre} dataset is grounded on a similar setting using the synthetic CLEVR universe~\cite{johnson2017clevr}. To measure causal induction beyond covariation, we challenge a visual reasoning system with $4$ types of queries in either independent scenarios or interventional scenarios: direct, indirect, screening-off, and backward-blocking. The first $2$ types of queries can be answered by counting co-occurrence, while the last $2$ types require in-depth causal representation. To better measure generalization in causal discovery, we further propose the compositionality and the systematicity \ac{ood} split.

We devise an optimization-based neuro-symbolic method to equip a visual reasoning system with the causal discovery ability. In particular, we decompose the model into a neural perception frontend and a causal reasoning backend. The neural perception frontend parses a given trial using a Mask RCNN~\cite{he2017mask}, whereas the causal reasoning backend performs continuous optimization for causal discovery~\cite{zheng2018dags,zheng2020learning}. The context trials are leveraged to learn a generalized \ac{sem}, and the answer to a query trial is solved by finding the best value to fit the \ac{sem}. As the first attempt, we separately train the two components, leaving the problem of closing the loop between visual perception and causal discovery for future work~\cite{li2020closed,zhang2021abstract,zhang2019metastyle}.

Existing visual reasoning systems' causal induction capability has been benchmarked on \ac{acre}. Specifically, we notice that pure neural models tend to perform causal reasoning by capturing the statistical correlation, achieving satisfactory results on direct and indirect queries but failing on screening-off and backward-blocking ones. For neuro-symbolic models, we notice that all of them struggle on backward-blocking and that the sparse and limited observation further adds to the complexity of the problem. Comparing performances of these $2$ types of models on various queries, we hypothesize that further combining learning and symbolic reasoning would be a promising direction for causal induction and broader causal reasoning problems.

At length, we hope challenges in this causal reasoning task would call for attention into visual systems with human-level spatial, temporal, and causal reasoning ability.

\noindent\textbf{Acknowledgement:}
The work reported herein was supported by ONR MURI N00014-16-1-2007, DARPA XAI N66001-17-2-4029, and ONR N00014-19-1-2153.

\small
\bibliographystyle{ieee_fullname}
\bibliography{bib}

\begin{thebibliography}{10}\itemsep=-1pt

\bibitem{baradel2020cophy}
Fabien Baradel, Natalia Neverova, Julien Mille, Greg Mori, and Christian Wolf.
\newblock Cophy: Counterfactual learning of physical dynamics.
\newblock In {\em International Conference on Learning Representations (ICLR)},
  2020.

\bibitem{battaglia2013simulation}
Peter~W Battaglia, Jessica~B Hamrick, and Joshua~B Tenenbaum.
\newblock Simulation as an engine of physical scene understanding.
\newblock {\em Proceedings of the National Academy of Sciences (PNAS)},
  110(45):18327--18332, 2013.

\bibitem{bramley2018intuitive}
Neil~R Bramley, Tobias Gerstenberg, Joshua~B Tenenbaum, and Todd~M Gureckis.
\newblock Intuitive experimentation in the physical world.
\newblock {\em Cognitive Psychology}, 105:9--38, 2018.

\bibitem{byrd1995limited}
Richard~H Byrd, Peihuang Lu, Jorge Nocedal, and Ciyou Zhu.
\newblock A limited memory algorithm for bound constrained optimization.
\newblock {\em SIAM Journal on scientific computing}, 16(5):1190--1208, 1995.

\bibitem{carpenter1990one}
Patricia~A Carpenter, Marcel~A Just, and Peter Shell.
\newblock What one intelligence test measures: a theoretical account of the
  processing in the raven progressive matrices test.
\newblock {\em Psychological review}, 97(3):404, 1990.

\bibitem{blender2016blender}
Blender~Online Community.
\newblock Blender--a 3d modelling and rendering package, 2016.

\bibitem{cook2011science}
Claire Cook, Noah~D Goodman, and Laura~E Schulz.
\newblock Where science starts: Spontaneous experiments in preschoolers’
  exploratory play.
\newblock {\em Cognition}, 120(3):341--349, 2011.

\bibitem{devlin2019bert}
Jacob Devlin, Ming-Wei Chang, Kenton Lee, and Kristina Toutanova.
\newblock Bert: Pre-training of deep bidirectional transformers for language
  understanding.
\newblock In {\em Proceedings of the Conference of the North American Chapter
  of the Association for Computational Linguistics (NAACL)}, 2019.

\bibitem{edmonds2019decomposing}
Mark Edmonds.
\newblock Decomposing human causal leanring: bottom-up associative learning and
  top-down schema reasoning.
\newblock In {\em Proceedings of the Annual Meeting of the Cognitive Science
  Society (CogSci)}, 2019.

\bibitem{edmonds2018human}
Mark Edmonds, James Kubricht, Colin Summers, Yixin Zhu, Brandon Rothrock,
  Song-Chun Zhu, and Hongjing Lu.
\newblock Human causal transfer: Challenges for deep reinforcement learning.
\newblock In {\em Proceedings of the Annual Meeting of the Cognitive Science
  Society (CogSci)}, 2018.

\bibitem{edmonds2020theory}
Mark Edmonds, Xiaojian Ma, Siyuan Qi, Yixin Zhu, Hongjing Lu, and Song-Chun
  Zhu.
\newblock Theory-based causal transfer: Integrating instance-level induction
  and abstract-level structure learning.
\newblock In {\em Proceedings of AAAI Conference on Artificial Intelligence
  (AAAI)}, 2020.

\bibitem{fire2015learning}
Amy Fire and Song-Chun Zhu.
\newblock Learning perceptual causality from video.
\newblock {\em ACM Transactions on Intelligent Systems and Technology (TIST)},
  7(2):1--22, 2015.

\bibitem{fodor1988connectionism}
Jerry~A Fodor, Zenon~W Pylyshyn, et~al.
\newblock Connectionism and cognitive architecture: A critical analysis.
\newblock {\em Cognition}, 28(1-2):3--71, 1988.

\bibitem{frosch2012causal}
Caren~A Frosch, Teresa McCormack, David~A Lagnado, and Patrick Burns.
\newblock Are causal structure and intervention judgments inextricably linked?
  a developmental study.
\newblock {\em Cognitive Science}, 36(2):261--285, 2012.

\bibitem{gerstenberg2017eye}
Tobias Gerstenberg, Matthew~F Peterson, Noah~D Goodman, David~A Lagnado, and
  Joshua~B Tenenbaum.
\newblock Eye-tracking causality.
\newblock {\em Psychological science}, 28(12):1731--1744, 2017.

\bibitem{gerstenberg2017intuitive}
Tobias Gerstenberg and Joshua~B Tenenbaum.
\newblock Intuitive theories.
\newblock In {\em Oxford handbook of causal reasoning}, pages 515--548. Oxford
  University Press New York, NY, 2017.

\bibitem{girdhar2020cater}
Rohit Girdhar and Deva Ramanan.
\newblock Cater: A diagnostic dataset for compositional actions \& temporal
  reasoning.
\newblock In {\em International Conference on Learning Representations (ICLR)},
  2020.

\bibitem{glymour2019review}
Clark Glymour, Kun Zhang, and Peter Spirtes.
\newblock Review of causal discovery methods based on graphical models.
\newblock {\em Frontiers in genetics}, 10:524, 2019.

\bibitem{gopnik2012scientific}
Alison Gopnik.
\newblock Scientific thinking in young children: Theoretical advances,
  empirical research, and policy implications.
\newblock {\em Science}, 337(6102):1623--1627, 2012.

\bibitem{gopnik2004theory}
Alison Gopnik, Clark Glymour, David~M Sobel, Laura~E Schulz, Tamar Kushnir, and
  David Danks.
\newblock A theory of causal learning in children: causal maps and bayes nets.
\newblock {\em Psychological review}, 111(1):3, 2004.

\bibitem{gopnik2000detecting}
Alison Gopnik and David~M Sobel.
\newblock Detecting blickets: How young children use information about novel
  causal powers in categorization and induction.
\newblock {\em Child development}, 71(5):1205--1222, 2000.

\bibitem{gopnik2001causal}
Alison Gopnik, David~M Sobel, Laura~E Schulz, and Clark Glymour.
\newblock Causal learning mechanisms in very young children: Two-, three-, and
  four-year-olds infer causal relations from patterns of variation and
  covariation.
\newblock {\em Developmental psychology}, 37(5):620, 2001.

\bibitem{gopnik2012reconstructing}
Alison Gopnik and Henry~M Wellman.
\newblock Reconstructing constructivism: Causal models, bayesian learning
  mechanisms, and the theory theory.
\newblock {\em Psychological bulletin}, 138(6):1085, 2012.

\bibitem{gordon2019permutation}
Jonathan Gordon, David Lopez-Paz, Marco Baroni, and Diane Bouchacourt.
\newblock Permutation equivariant models for compositional generalization in
  language.
\newblock In {\em International Conference on Learning Representations (ICLR)},
  2019.

\bibitem{griffiths2009theory}
Thomas~L Griffiths and Joshua~B Tenenbaum.
\newblock Theory-based causal induction.
\newblock {\em Psychological Review}, 116(4):661, 2009.

\bibitem{han2019visual}
Chi Han, Jiayuan Mao, Chuang Gan, Josh Tenenbaum, and Jiajun Wu.
\newblock Visual concept-metaconcept learning.
\newblock In {\em Proceedings of Advances in Neural Information Processing
  Systems (NeurIPS)}, 2019.

\bibitem{he2017mask}
Kaiming He, Georgia Gkioxari, Piotr Doll{\'a}r, and Ross Girshick.
\newblock Mask r-cnn.
\newblock In {\em Proceedings of International Conference on Computer Vision
  (ICCV)}, 2017.

\bibitem{he2016deep}
Kaiming He, Xiangyu Zhang, Shaoqing Ren, and Jian Sun.
\newblock Deep residual learning for image recognition.
\newblock In {\em Proceedings of the IEEE Conference on Computer Vision and
  Pattern Recognition (CVPR)}, 2016.

\bibitem{hill2018learning}
Felix Hill, Adam Santoro, David Barrett, Ari Morcos, and Timothy Lillicrap.
\newblock Learning to make analogies by contrasting abstract relational
  structure.
\newblock In {\em International Conference on Learning Representations (ICLR)},
  2019.

\bibitem{hochreiter1997long}
Sepp Hochreiter and J{\"u}rgen Schmidhuber.
\newblock Long short-term memory.
\newblock {\em Neural computation}, 9(8):1735--1780, 1997.

\bibitem{hu2020hierarchical}
Sheng Hu, Yuqing Ma, Xianglong Liu, Yanlu Wei, and Shihao Bai.
\newblock Hierarchical rule induction network for abstract visual reasoning.
\newblock {\em arXiv preprint arXiv:2002.06838}, 2020.

\bibitem{johnson2017clevr}
Justin Johnson, Bharath Hariharan, Laurens van~der Maaten, Li Fei-Fei, C
  Lawrence~Zitnick, and Ross Girshick.
\newblock Clevr: A diagnostic dataset for compositional language and elementary
  visual reasoning.
\newblock In {\em Proceedings of the IEEE Conference on Computer Vision and
  Pattern Recognition (CVPR)}, 2017.

\bibitem{kingma2014adam}
Diederik~P Kingma and Jimmy Ba.
\newblock Adam: A method for stochastic optimization.
\newblock {\em arXiv preprint arXiv:1412.6980}, 2014.

\bibitem{kubricht2017intuitive}
James~R Kubricht, Keith~J Holyoak, and Hongjing Lu.
\newblock Intuitive physics: Current research and controversies.
\newblock {\em Trends in Cognitive Sciences}, 21(10):749--759, 2017.

\bibitem{kushnir2007conditional}
Tamar Kushnir and Alison Gopnik.
\newblock Conditional probability versus spatial contiguity in causal learning:
  Preschoolers use new contingency evidence to overcome prior spatial
  assumptions.
\newblock {\em Developmental psychology}, 43(1):186, 2007.

\bibitem{lagnado2007beyond}
David~A Lagnado, Michael~R Waldmann, York Hagmayer, and Steven~A Sloman.
\newblock Beyond covariation.
\newblock In {\em Causal learning: Psychology, philosophy, and computation},
  pages 154--172. Oxford University Press, 2007.

\bibitem{lake2018generalization}
Brenden Lake and Marco Baroni.
\newblock Generalization without systematicity: On the compositional skills of
  sequence-to-sequence recurrent networks.
\newblock In {\em Proceedings of International Conference on Machine Learning
  (ICML)}, 2018.

\bibitem{lake2017building}
Brenden~M Lake, Tomer~D Ullman, Joshua~B Tenenbaum, and Samuel~J Gershman.
\newblock Building machines that learn and think like people.
\newblock {\em Behavioral and Brain Sciences}, 40, 2017.

\bibitem{li2020closed}
Qing Li, Siyuan Huang, Yining Hong, Yixin Chen, Ying~Nian Wu, and Song-Chun
  Zhu.
\newblock Closed loop neural-symbolic learning via integrating neural
  perception, grammar parsing, and symbolic reasoning.
\newblock {\em arXiv preprint arXiv:2006.06649}, 2020.

\bibitem{lin2017feature}
Tsung-Yi Lin, Piotr Doll{\'a}r, Ross Girshick, Kaiming He, Bharath Hariharan,
  and Serge Belongie.
\newblock Feature pyramid networks for object detection.
\newblock In {\em Proceedings of the IEEE Conference on Computer Vision and
  Pattern Recognition (CVPR)}, 2017.

\bibitem{lopez2017discovering}
David Lopez-Paz, Robert Nishihara, Soumith Chintala, Bernhard Scholkopf, and
  L{\'e}on Bottou.
\newblock Discovering causal signals in images.
\newblock In {\em Proceedings of the IEEE Conference on Computer Vision and
  Pattern Recognition (CVPR)}, 2017.

\bibitem{lucas2014children}
Christopher~G Lucas, Sophie Bridgers, Thomas~L Griffiths, and Alison Gopnik.
\newblock When children are better (or at least more open-minded) learners than
  adults: Developmental differences in learning the forms of causal
  relationships.
\newblock {\em Cognition}, 131(2):284--299, 2014.

\bibitem{mao2019neuro}
Jiayuan Mao, Chuang Gan, Pushmeet Kohli, Joshua~B Tenenbaum, and Jiajun Wu.
\newblock The neuro-symbolic concept learner: Interpreting scenes, words, and
  sentences from natural supervision.
\newblock In {\em International Conference on Learning Representations (ICLR)},
  2019.

\bibitem{meltzoff2012learning}
Andrew~N Meltzoff, Anna Waismeyer, and Alison Gopnik.
\newblock Learning about causes from people: Observational causal learning in
  24-month-old infants.
\newblock {\em Developmental psychology}, 48(5):1215, 2012.

\bibitem{michotte2017perception}
Albert Michotte.
\newblock {\em The perception of causality}, volume~21.
\newblock Routledge, 2017.

\bibitem{mottaghi2016happens}
Roozbeh Mottaghi, Mohammad Rastegari, Abhinav Gupta, and Ali Farhadi.
\newblock “what happens if...” learning to predict the effect of forces in
  images.
\newblock In {\em Proceedings of European Conference on Computer Vision
  (ECCV)}, 2016.

\bibitem{paszke2017automatic}
Adam Paszke, Sam Gross, Soumith Chintala, Gregory Chanan, Edward Yang, Zachary
  DeVito, Zeming Lin, Alban Desmaison, Luca Antiga, and Adam Lerer.
\newblock Automatic differentiation in pytorch.
\newblock {\em NIPS-W}, 2017.

\bibitem{pearl2009causality}
Judea Pearl.
\newblock {\em Causality}.
\newblock Cambridge university press, 2009.

\bibitem{pearl1995theory}
Judea Pearl and Thomas~S Verma.
\newblock A theory of inferred causation.
\newblock In {\em Studies in Logic and the Foundations of Mathematics}, volume
  134, pages 789--811. Elsevier, 1995.

\bibitem{qi2020generalized}
Siyuan Qi, Baoxiong Jia, Siyuan Huang, Ping Wei, and Song-Chun Zhu.
\newblock A generalized earley parser for human activity parsing and
  prediction.
\newblock {\em IEEE Transactions on Pattern Analysis and Machine Intelligence
  (TPAMI)}, 2020.

\bibitem{qi2018generalized}
Siyuan Qi, Baoxiong Jia, and Song-Chun Zhu.
\newblock Generalized earley parser: Bridging symbolic grammars and sequence
  data for future prediction.
\newblock In {\em Proceedings of International Conference on Machine Learning
  (ICML)}, 2018.

\bibitem{raven1938raven}
J.~C. et~al. Raven.
\newblock Raven’s progressive matrices.
\newblock {\em Western Psychological Services}, 1938.

\bibitem{rescorla1972theory}
Robert~A Rescorla.
\newblock A theory of pavlovian conditioning: Variations in the effectiveness
  of reinforcement and nonreinforcement.
\newblock {\em Current research and theory}, pages 64--99, 1972.

\bibitem{ruis2020benchmark}
Laura Ruis, Jacob Andreas, Marco Baroni, Diane Bouchacourt, and Brenden~M Lake.
\newblock A benchmark for systematic generalization in grounded language
  understanding.
\newblock {\em arXiv preprint arXiv:2003.05161}, 2020.

\bibitem{santoro2018measuring}
Adam Santoro, Felix Hill, David Barrett, Ari Morcos, and Timothy Lillicrap.
\newblock Measuring abstract reasoning in neural networks.
\newblock In {\em Proceedings of International Conference on Machine Learning
  (ICML)}, 2018.

\bibitem{santoro2017simple}
Adam Santoro, David Raposo, David~G Barrett, Mateusz Malinowski, Razvan
  Pascanu, Peter Battaglia, and Timothy Lillicrap.
\newblock A simple neural network module for relational reasoning.
\newblock In {\em Proceedings of Advances in Neural Information Processing
  Systems (NeurIPS)}, 2017.

\bibitem{scholl2000perceptual}
Brian~J Scholl and Patrice~D Tremoulet.
\newblock Perceptual causality and animacy.
\newblock {\em Trends in Cognitive Sciences}, 4(8):299--309, 2000.

\bibitem{schulz2007learning}
Laura Schulz, Tamar Kushnir, and Alison Gopnik.
\newblock Learning from doing: Intervention and causal inference.
\newblock {\em Causal learning: Psychology, philosophy, and computation}, pages
  67--85, 2007.

\bibitem{shanks1988associative}
David~R Shanks and Anthony Dickinson.
\newblock Associative accounts of causality judgment.
\newblock In {\em Psychology of learning and motivation}, volume~21, pages
  229--261. Elsevier, 1988.

\bibitem{sobel2006blickets}
David~M Sobel and Natasha~Z Kirkham.
\newblock Blickets and babies: the development of causal reasoning in toddlers
  and infants.
\newblock {\em Developmental psychology}, 42(6):1103, 2006.

\bibitem{sobel2004children}
David~M Sobel, Joshua~B Tenenbaum, and Alison Gopnik.
\newblock Children's causal inferences from indirect evidence: Backwards
  blocking and bayesian reasoning in preschoolers.
\newblock {\em Cognitive science}, 28(3):303--333, 2004.

\bibitem{spirtes2000causation}
Peter Spirtes, Clark~N Glymour, Richard Scheines, and David Heckerman.
\newblock {\em Causation, prediction, and search}.
\newblock MIT press, 2000.

\bibitem{spratley2020closer}
Steven Spratley, Krista Ehinger, and Tim Miller.
\newblock A closer look at generalisation in raven.
\newblock In {\em Proceedings of European Conference on Computer Vision
  (ECCV)}, 2020.

\bibitem{twain1984life}
Mark Twain.
\newblock {\em Life on the Mississippi}.
\newblock Penguin, 1984.

\bibitem{waldmann1992predictive}
Michael~R Waldmann and Keith~J Holyoak.
\newblock Predictive and diagnostic learning within causal models: Asymmetries
  in cue competition.
\newblock {\em Journal of Experimental Psychology: General}, 121(2):222, 1992.

\bibitem{walker2014toddlers}
Caren~M Walker and Alison Gopnik.
\newblock Toddlers infer higher-order relational principles in causal learning.
\newblock {\em Psychological science}, 25(1):161--169, 2014.

\bibitem{walker2017discriminating}
Caren~M Walker and Alison Gopnik.
\newblock Discriminating relational and perceptual judgments: Evidence from
  human toddlers.
\newblock {\em Cognition}, 166:23--27, 2017.

\bibitem{wang2020abstract}
Duo Wang, Mateja Jamnik, and Pietro Lio.
\newblock Abstract diagrammatic reasoning with multiplex graph networks.
\newblock In {\em International Conference on Learning Representations (ICLR)},
  2020.

\bibitem{xie2021halma}
Sirui Xie, Xiaojian Ma, Peiyu Yu, Yixin Zhu, Ying~Nian Wu, and Song-Chun Zhu.
\newblock Halma: Humanlike abstraction learning meets affordance in rapid
  problem solving.
\newblock {\em arXiv preprint arXiv:2102.11344}, 2021.

\bibitem{yi2020clevrer}
Kexin Yi, Chuang Gan, Yunzhu Li, Pushmeet Kohli, Jiajun Wu, Antonio Torralba,
  and Joshua~B Tenenbaum.
\newblock Clevrer: Collision events for video representation and reasoning.
\newblock In {\em International Conference on Learning Representations (ICLR)},
  2020.

\bibitem{yi2018neural}
Kexin Yi, Jiajun Wu, Chuang Gan, Antonio Torralba, Pushmeet Kohli, and Josh
  Tenenbaum.
\newblock Neural-symbolic vqa: Disentangling reasoning from vision and language
  understanding.
\newblock In {\em Proceedings of Advances in Neural Information Processing
  Systems (NeurIPS)}, 2018.

\bibitem{zhang2019raven}
Chi Zhang, Feng Gao, Baoxiong Jia, Yixin Zhu, and Song-Chun Zhu.
\newblock Raven: A dataset for relational and analogical visual reasoning.
\newblock In {\em Proceedings of the IEEE Conference on Computer Vision and
  Pattern Recognition (CVPR)}, 2019.

\bibitem{zhang2019learning}
Chi Zhang, Baoxiong Jia, Feng Gao, Yixin Zhu, Hongjing Lu, and Song-Chun Zhu.
\newblock Learning perceptual inference by contrasting.
\newblock In {\em Proceedings of Advances in Neural Information Processing
  Systems (NeurIPS)}, 2019.

\bibitem{zhang2021abstract}
Chi Zhang, Baoxiong Jia, Song-Chun Zhu, and Yixin Zhu.
\newblock Abstract spatial-temporal reasoning via probabilistic abduction and
  execution.
\newblock In {\em Proceedings of the IEEE Conference on Computer Vision and
  Pattern Recognition (CVPR)}, 2021.

\bibitem{zhang2019metastyle}
Chi Zhang, Yixin Zhu, and Song-Chun Zhu.
\newblock Metastyle: Three-way trade-off among speed, flexibility, and quality
  in neural style transfer.
\newblock In {\em Proceedings of AAAI Conference on Artificial Intelligence
  (AAAI)}, 2019.

\bibitem{zhang2020machine}
Wenhe Zhang, Chi Zhang, Yixin Zhu, and Song-Chun Zhu.
\newblock Machine number sense: A dataset of visual arithmetic problems for
  abstract and relational reasoning.
\newblock In {\em Proceedings of AAAI Conference on Artificial Intelligence
  (AAAI)}, 2020.

\bibitem{zheng2019abstract}
Kecheng Zheng, Zheng-Jun Zha, and Wei Wei.
\newblock Abstract reasoning with distracting features.
\newblock In {\em Proceedings of Advances in Neural Information Processing
  Systems (NeurIPS)}, 2019.

\bibitem{zheng2018dags}
Xun Zheng, Bryon Aragam, Pradeep~K Ravikumar, and Eric~P Xing.
\newblock Dags with no tears: Continuous optimization for structure learning.
\newblock In {\em Proceedings of Advances in Neural Information Processing
  Systems (NeurIPS)}, 2018.

\bibitem{zheng2020learning}
Xun Zheng, Chen Dan, Bryon Aragam, Pradeep Ravikumar, and Eric Xing.
\newblock Learning sparse nonparametric dags.
\newblock In {\em Proceedings of the International Conference on Artificial
  Intelligence and Statistics (AISTATS)}, 2020.

\bibitem{zhu2020dark}
Yixin Zhu, Tao Gao, Lifeng Fan, Siyuan Huang, Mark Edmonds, Hangxin Liu, Feng
  Gao, Chi Zhang, Siyuan Qi, Ying~Nian Wu, Josh~B Tenenbaum, and Song-Chun Zhu.
\newblock Dark, beyond deep: A paradigm shift to cognitive ai with humanlike
  common sense.
\newblock {\em Engineering}, 6(3):310--345, 2020.

\end{thebibliography}

\end{document}